\documentclass{article}
\usepackage{ijcai16}

\usepackage{times}

\usepackage{algorithm}
\usepackage[noend]{algpseudocode}
\usepackage{graphicx} \usepackage{multirow}

  \newtheorem{definition}{Definition}

\newcommand{\seq}[1]{\langle #1 \rangle}

\begin{document}

\title{SLIM: Semi-Lazy Inference Mechanism for Plan Recognition}
\author{Reuth Mirsky \and Ya'akov (Kobi) Gal \\
        Department of Information Systems Engineering \\ Ben-Gurion
  University of the Negev, Israel \\ \{dekelr@post.bgu.ac.il,kobig@bgu.ac.il\}}

\maketitle

\begin{abstract}
Plan Recognition algorithms require to recognize a complete  hierarchy explaining the agent's actions and goals. While the output of such algorithms is informative to the recognizer, the cost of its calculation is high in run-time, space, and completeness. Moreover, performing plan recognition online requires the observing agent to reason about future actions that have not yet been seen and maintain a set of hypotheses to support all possible options.
This paper presents a new and efficient algorithm for online plan recognition called SLIM (Semi-Lazy Inference Mechanism). It combines both a bottom-up and top-down parsing processes, which allow it to commit only to the minimum necessary actions in real-time, but still provide complete hypotheses post factum.
We show both theoretically and empirically that although the computational cost of this process is still exponential, there is a significant improvement in run-time when compared to a state of the art of plan recognition algorithm.
\end{abstract}

\section{Introduction}
As intelligent systems become prevalent in our daily lives, our interactions can benefit greatly when these systems are able to  understand our activities.
This recognition problem can be roughly divided into two types of
tasks: goal recognition and plan recognition. The goal recognition task  can be viewed as a classification problem, where we receive a sequence of actions and need to label this sequence with a goal. For example, consider a case where we observe someone's actions in the kitchen. Depending on the granularity of our goals, we can recognize ``baking'' ``baking a cake'', ``baking a chocolate cake'', etc.  

The plan recognition task, on the other hand, focuses on the complete hierarchy and dependencies between the observed actions. Given a sequence of observations, the plan recognition task is to provide with either a hierarchy of activities that explains the observations. Using the same kitchen example, a potential agent who performs plan recognition can hypothesize in real time about how the person is going to make a chocolate cake. This agent will potentially be able to assist the cook or prevent  failures, while a goal recognition agent can only change the label from ``baking a chocolate cake'' to ``failing to bake a chocolate cake''.

This motivation can be seen in various domains: assist disaster response crews by recognizing the protocol they are using~\cite{blaylock2006fast}, advise about missing steps in the medical guideline a doctor is following~\cite{charniak2013plan,ng1992abductive}, assist to students learning via online software~\cite{AG13,uzan2015plan} and provide real-time response to a cyber attack according to its specific actions~\cite{bisson2011provoking,qin2004attack}.

The plan recognition task is more costly than the goal
recognition task.  This is because it must output a complete
hierarchy of plans that explain the observed actions.  Existing online
plan recognition algorithms can only process small sequences of
actions,  work on  limited domains, or hinder the completeness of
their outputted hypotheses ~\cite{wiseman2014discriminatively,kabanza2013controlling}. 
Consider
the baking scenario, where the cook starts with pouring flour and
sugar - many cakes start with these steps, so keeping all possible
hypotheses that explain the cook's actions will require a plan recognition
algorithm to generate many different plans. In general, the number
of hypotheses may grow very large even for a small number of
observations~\cite{spr2016}.

The main contribution of this paper is a new type of plan recognition
algorithm that combines both bottom-up and top-down parsing processes
 called SLIM (Semi-Lazy Inference Mechanism).  
Rather than maintain complete plan hierarchies, SLIM constructs plans from the bottom up, according to a  least commitment policy which adds the  smallest plan fragments that explain the observations. 
We present a theoretical analysis showing that SLIM requires to consider less combinations than the state of the art.  We then show empirically that the description of local behavior is easier to maintain in real-time and leads to significantly faster performances on a domain from the plan recognition literature~\cite{kabanza2013controlling}.

\section {Related Work}
 Bui~\shortcite{bui2003general} use particle filtering to
provide approximate solutions to plan recognition problems.
Ramirez and
Geffner~\shortcite{ramirez2009plan} use a planning domain definition which allows an implicit representation of the possible plans.
Sukthankar and Sycara~\shortcite{sukthankar2008hypothesis} propose
heuristics (based on temporal constraints) to narrow the number of
hypotheses to consider during recognition but require that the full
plan library be generated in advance. Other works ~\cite{conati1997line,katz2007out} use plan recognition algorithms to infer
students' plans to solve problems in a simulated physics environment
by comparing their actions to a set of predefined possible plans. 
None of these works was compared to each other and each was evaluated on domain-specific problems. 

Shieber and Wiseman~\shortcite{wiseman2014discriminatively} use a
discriminative re-ranking approach in abductive settings in order to
infer plans on the same data set; their work requires a training set
and is not appropriate for online use. Other works which focused on the offline case of plan recognition, where the inference is performed after the agent completed execution~\cite{AG13,uzan2015plan}. However, these works cannot be of use if the recognition is needed in real-time.

Geib and Goldman presented two probabilistic online recognition
 algorithms -- The first is PHATT~\cite{GeibGoldman09}, that builds all possible plans incrementally with each new observation. 
This approach maintains all possible hypotheses for matching future
unseen actions by the agent. While this approach is complete,
it is very slow. The second, YAPPR~\cite{geib2008new} was devised
to be more efficient, but at the cost of symbolic representation of the
plans. This approach  outputs  a  distribution
over the agent's inferred goals. 
  Kabanza et al.~\shortcite{kabanza2013controlling} extended this approach to
compute lower and upper bounds on the goal hypothesis probabilities
and prune the search space of possible plans. Again, this algorithm only outputs
a distribution over the set of goals.
Another work by Geib~\shortcite{geib2009delaying} suggests the use of combinatory categorial grammars (CCGs). This representation is more compact at the cost of  expressiveness, as it outputs probabilities on different categories instead of complete plans.
 
Avrahami and Kaminka's SBR~\shortcite{AvrahamiKaminka05} suggested intelligent tree-based representation for plan recognition and an algorithm that traverses the trees and log actions in a manner that is temporally consistent with the observations. Upon query request, the logged actions can be used to construct the possible hypotheses at the time of the query. This delayed construction causes the algorithm to perform only minimal commitments about matching actions to the grammar.  This work was extended by Fagundes et al.~\shortcite{fagundes2014dealing} to provide an anytime expectation of time needed to recognize the plan, and by Avrahami and Kaminka ~\shortcite{avrahami2007incorporating} to reason about the cost of 
The late commitment of SBR makes it
extremely efficient in terms of time, but at the cost of space. In order to be able
to provide with a hypothesis when needed, the algorithm must explicitly generate
all paths that were possibly taken. SLIM was inspired by both PHATT's incremental
approach and by SBR's lazy commitment.

Other previous works tried to reduce runtime complexity of the plan recognition process by giving up on the more specific details of the plan or hindering completeness~\cite{geib2009delaying,geib2008new}. 

\section{Background}
We follow the definition by Geib and Goldman, in which we assume
 that the plan recognition algorithm is given a plan library which
can implicitly describe any expected behaviors of the observed agent.
We simplified their definition for clarity.
\begin{definition} [Plan Library] A plan library is a tuple $L=\langle \Sigma,NT,G,R\rangle$, 
where $\Sigma$ is a finite set of terminal letters, $NT$ is a finite set of non-terminal letters, $G \subset NT$ is the goal actions and $R$ is a set of production rules of the form $A \rightarrow \alpha : \phi, p$, where:
\begin {enumerate}
\item $A \in NT$
\item $\alpha$ is a string of symbols from $(\Sigma \cup NT)*$
\item $\phi = \{ (i,j) \mid \alpha[i] \leq \alpha[j] \}$ where $\alpha[i]$
 and $\alpha[j]$ refer to the i-th and j-th symbols in $\alpha$, respectively.
\item $p$ is the prior probability for using this rule to execute $A$. 
\end{enumerate}
\label{def:plan-library}
\end{definition}

This definition follows traditional PCFG based encoding for hierarchical plans~\cite{PynadathWellman00}. This  representation is also similar to fundamental planning formalisms such as Hierarchical Task Networks (HTNs)~\cite{ghallab2004automated} but does not include  action preconditions and effects.
$\Sigma$ represents the basic actions that can be observed. $NT$ represents complex actions, which are different levels of abstract actions that can be expanded into sequences of both basic and complex actions. $G$ is a subset of $NT$, representing the highest level of abstraction to describe a plan -- by the goal it tries to accomplish. Each $r \in R$ is a production rule, describing how a complex action (the left side of the rule) can be accomplished by a sequence of both basic and complex actions (the right side of the rule). 
The tuples described in $\phi$ are explicit ordering constraints, similar to ID/LP grammars ~\cite{Shieber84}. 
A {\em plan} for achieving a complex action $c\in NT$ is a tree whose
root is labeled by $c$, and each parent node is labeled with a complex
action such that its children nodes are a decomposition of its complex
action into constituent actions according to one of the production rules in $R$.
The temporal constraints of the production rules enforce the order that the 
actions can be observed. In this paper, we use the terms ``plan'' and ``tree'' interchangeably.

A plan is said to be {\em complete} if all its leaf nodes are labeled
with basic actions.  Incomplete plans include complex level actions
that have not been decomposed using a production rule. These
\emph{open frontier} nodes represent actions that are expected to
be carried out by the agent in the future~\cite{GeibGoldman09}.

\begin{figure}[t]
 \centering
	\includegraphics[width=7cm]{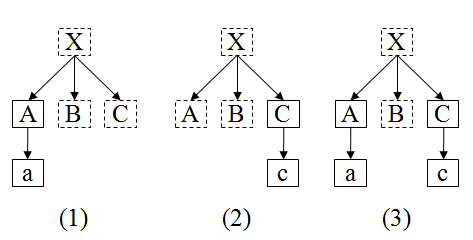}
 \caption{Example Plans}
        \label{fig:example}
 \end{figure}

For example, consider the following plan library:
\begin {description}
\item [Basic actions] $\Sigma=\{a,b,c\}$
 \item [Complex actions] $NT=\{X, A, B, C\}$
\item [Goals] $G=\{X\}$ 
\item [Production rules] Three rules for A,B,C of the form: $A \rightarrow a \mid \emptyset$ and a rule for achieving the goal $X \rightarrow A, B, C \mid \{(1,2)\}$. The ordering constraint in this rule means that A must be achieved before B.
\end{description}
Each of the three plans in Figure~\ref{fig:example} are incomplete plans for achieving the complex action $X$. Nodes with dashed outline represent actions that have not yet been realized. Leaves with dashed outline represent open frontier nodes. As expected, the nodes in valid plans are always consistent with the rule's ordering constraints $\phi$. This means that these plans are valid only if there isn't any constraint  $(i,j)$ in the rule if the i-th node is dashed and the j-th is not. Notice that the node for $X$ is dashed as well, since in neither of these plans was $X$ fully achieved yet. 

An \emph{observation sequence} is an ordered set of basic actions that
represents actions carried out by the observed agent.  A plan
\emph{describes} an observation sequence if each observation is mapped
to a leaf node in the plan. The observed agent is assumed to plan by
choosing a subset of complex actions as intended goals and then
carrying out a separate plan for completing each of these goals.
Using the example from Figure~\ref{fig:example}, plan (a) explains the observation $a$, plan (b) explains $c$ and plan (c) explains $a,c$.

An agent may execute more than one plan intermittently, and the observations explained by a given plan may not be contiguous. 
A  \emph{hypothesis} for a given observation sequence is a  set of plans.
 Each of the plans in a hypothesis satisfies one of the goals that the agent is
carrying out. 
For example, a possible hypothesis for the observation sequence $a,c$ can be plan (c) in Figure~\ref{fig:example}, while another hypothesis can be a set of plans (a) and (b), which together explain the sequence $a,c$.

We assume that each hypothesis is assigned a  probability representing the belief that  this is the agent's intended set of plans, given the set of observations. This probability is calculated using the production rules' probabilities, in a similar fashion to Geib and Goldman~\shortcite{GeibGoldman09}.

Both PHATT and the newly presented SLIM perform incrementally -- at each step $n$, the algorithm gets a new observation $\sigma_n$ and tries to combine it with the hypotheses that explain $\sigma_1, \ldots, \sigma_{n-1}$. The difference between the algorithms is in the construction process.

At the first observation $\sigma_0$, PHATT constructs a partial tree
where the root is labeled with a goal  $g\in G$ and a path in the tree from $G$ to
the observation. It is required that all leaves in the tree 
are open frontier nodes labeled with non-terminal symbols in $NT$,  except the leaf representing the  observation itself 
which is a terminal symbol in $\Sigma$.  This leaf is called the \emph{leftmost child} of the sub-tree, 
while the partial tree is referred to as a \emph{leftmost tree}. 
The  leftmost child represent the next possible action that can be added to the plan. 
 PHATT attempts to combine each new observation
$\sigma_i$ to each hypothesis in the set of previous hypotheses,
either by connecting the partial tree to  an existing plan, or as a new plan in the
hypothesis.
We  formalize this intuition using the following definitions:
\begin{definition}[leftmost child]
  Let $T$ be a plan and let $\alpha_0$ be an internal node
  in $T$ with children
  $\seq{\alpha_1, \ldots, \alpha_{i-1},    \alpha_i,  \alpha_{i+1} \ldots,   \alpha_n}$.
  We say that $\alpha_i$ is a \emph{left-most child} of node
  $\alpha_0$ if:
\begin {enumerate}
\item $\alpha_i$ is an open frontier in $T$.
\item $\exists r\in R \mid \alpha_0 \rightarrow \alpha_1, \ldots, \alpha_{i-1}, \alpha_i,  \alpha_{i+1} \ldots, \alpha_n : \phi$ such that for every child node labeled $\alpha_j,$ such that $j \neq i$, it holds that $\alpha_j$ is not in the open frontier of $T$.
\end{enumerate}
\end{definition}
\begin{definition}[leftmost tree]
  A \emph{leftmost tree} deriving an observation
  $\sigma_t$ is a plan with frontier
  $\seq{\alpha_1, \ldots, \sigma_i,\ldots,
    \alpha_n}$ such that
  \begin {enumerate}
  \item $\sigma_i\in\Sigma$ and $\alpha_j \in NT$ for all $j \neq i$;
    and
  \item For any child node labeled $\alpha_j$ and its parent
    $\alpha_k$ in the path from $\sigma_i$
    to the root, the node $\alpha_j$ is a leftmost
    child of $\alpha_k$.
  \end{enumerate}
\end{definition}

In Figure~\ref{fig:example}, plan (1) has two leftmost children which
are $B, C$, plan (2) has one leftmost child which is $A$ and plan (3)
has one leftmost child which is $B$.  Notice that although there are
two nodes in the open frontier  in plan (2), only the leftmost child  $A$ can be considered for expanding this plan.
Only plans (1) and (2) are leftmost trees, since plan (3) has two
leftmost children.  

When introducing $\sigma_i$ into an existing hypothesis, PHATT considers two possibilities:
\begin{enumerate}
\item Adding the observation as a new plan in the hypothesis, which
  requires to construct a leftmost tree that derives $\sigma_i$ with a
  root in $G$, the set of goals.
\item Updating an  existing plan in the hypothesis with the observation, which requires to  construct a leftmost tree deriving $\sigma_i$ with a root which matches one of the open frontier nodes of an  existing plan.
\end{enumerate}

A disadvantage of this process is that each plan is explicitly
represented in the hypothesis and the algorithm must maintain, for each hypothesis in the set, all possible plans that explain the observations seen thus far. Even for a small number of observations, the size of a plan  in the corresponding hypothesis may be very large. For example, if the baker is taking out eggs, the algorithm should recognize that the next step is finding the whisk, no matter which type of cake is being prepared. However, PHATT must keep a separate hypothesis for each possible recipe - from chocolate cake to red velvet - and keep those deep plans through all next steps.

In order to reduce this cost, we propose that the   incremental construction of a 
plan does not  
 commit to a specific path between  a goal and an observation until it is specifically needed.  

\section {The SLIM Algorithm}

The motivation for SLIM derives from the need to infer the agent's plans in real-time. In many cases, the goal the agent is pursuing is less important, but rather how the observed actions combine together locally. Consider a known use case from the cyber security domain of an advanced persistent threat (APT) -- an agent wishes to take use of resources and exploit weaknesses over time. We wish to use plan recognition to detect and response to such attacks -- at any given point in time, we do not need to know exactly what attack is being devised on the long run, but rather which local protocols are used (identity theft, permissions sharing, etc.) in order to respond to them specifically. In such scenarios, PHATT's production of complete paths to the root causes unnecessary overhead. SLIM works similarly to PHATT,  but instead of producing a 
complete path to a possible root for each observation, it simply creates a 1-depth tree with the observation as a leaf. Each such tree is called a fragment.
\begin{definition}
[fragment] A \emph{fragment} for an action $\sigma$ is a leftmost tree deriving $\sigma$ with a depth of 1.
\end{definition}

Given a set of fragments for observation $\sigma_n$, SLIM tries to combine each new fragment with the hypothesis $h \in H_{n-1}$, either by adding it as a standalone fragment (the function $\Call{CombineIndependently}{}$) or by fusing it with other plans (the functions $\Call{CombineAsChild}{}$ and $\Call{CombineAsSibling}{}$).

\begin{definition}
[Fusion] A plan $p$ can be \emph{fused} with a fragment $f$ if (1) there exists a node $o$ in the plan's open frontier; (2) the root of $f$ is the same action as $o$; (3) there are no ordering constraints preventing from $o$ to be executed next.
\end{definition}

For example, consider plan $2$ from Figure~\ref{fig:example} -- although both $A$ and $B$ are in the plan's open frontier, a fusion can be performed only with a fragment whose root node is $A$, since $B$ must be executed after $A$.
We extend the definition of fusion to a hypothesis and fragment, where the one of the plans in the hypothesis can be fused with a fragment. A hypothesis in SLIM is a set of fused fragments which describe the observations. A \emph{local hypothesis} is a set of fused fragments.

The SLIM algorithm includes two processes, one that constructs local hypotheses  from the bottom-up, 
and one that constructs complete hypotheses using a top-down approach. 
The resulting improvement of SLIM in comparison to PHATT, is that it does not need to carry long paths to the root unless it is specifically asked for by the top-down process. It is similar to the SBR algorithm~\cite{AvrahamiKaminka05}, which only keeps the ``current state'' unless it is required to complete the paths. However, it does commit (in each hypothesis) to one small fragment of the plan, thus making it only a semi-lazy algorithm, where SBR is defined lazy and does not commit to a path until it is necessary.
In order to make sure that all ordering constraints are preserved, each fragment also keeps a timestamp, in a similar way to SBR. This timestamp will help in deciding whether two fragments that were developed individually can be combined in such a way that the ordering constrains will not be damaged.  

\begin{figure}[t]
\small
\begin{algorithmic}
\Function {SLIM Bottom-Up} {$H_{n-1}, \sigma_n$} 
	\State $H_n \gets H_n \cup \Call{CombineDirectly}{H_{n-1}, \sigma_n}$
	\State $F_n \gets \Call{createFragments}{\sigma_n}$
	\ForAll{f $ \in F_n $}\
		\State $H_n \gets H_n \cup \Call{CombineAsChild}{H_{n-1}, f}$
		\State $H_n \gets H_n \cup \Call{CombineAsSibling}{H_{n-1}, f}$
		\State $H_n \gets H_n \cup \Call{CombineIndependently}{H_{n-1}, f}$
	\EndFor
	\State \textbf{Return} $H_n$
\EndFunction

\Statex

\Function {CombineDirectly} {$H_{n-1}, \sigma_n$} 
	\ForAll{$h \in H_{n-1}$}
		\ForAll{open frontier node  $o \in h$ }
			\State $h' \gets \Call{Fuse}{h, o, \sigma_n}$
			\If {h' is a valid hypothesis}
	                               	\State $H_n \gets H_n\cup h'$
			\EndIf
		\EndFor
	\EndFor
\EndFunction

\Statex

%
%

\Function {CombineAsSibling} {$H_{n-1}, f$} 
	\State $H_n \gets \emptyset$
	\ForAll{$h \in H_{n-1}$}
		\ForAll{plan $p\in h$}
			\State $F' \gets \Call{createFragments}{f.root()}$
			\ForAll {$f'\in F'$ describing $p$ and $f$ as nodes $\alpha_i, \alpha_j$}
				\State $p' \gets \Call{Fuse}{f', \alpha_i, p} $
				\State $p' \gets \Call{Fuse}{f', \alpha_j, f} $
				\State $h' \gets h \backslash p \cup p'$
				\If {h' is a valid hypothesis}
	         			          	\State $H_n \gets H_n\cup h'$
				\EndIf
			\EndFor
		\EndFor
	\EndFor

	\Statex
	\State \textbf{Return} $H_n$
\EndFunction

%

\end{algorithmic}  
\label{fig:algs}
\caption{SLIM's Main Functions}
\end{figure}

\subsection {Bottom-Up Inference}
A main contribution of the SLIM is its process of deciding when two fragments are related and can be fused into one plan. After seeing $\sigma_1, \ldots, \sigma_{n-1}$, SLIM holds a local hypothesis set $H_{n-1}$ for describing the observations so far. When seeing a new observation, $\sigma_n$,  SLIM calls \Call{CreateFragments}{} to generate a set of fragments for $\sigma_n$.
Then SLIM tries to combine each of the fragments to each of the hypotheses $h \in H_{n-1}$ using each of the following approaches:
\begin {enumerate}
\item As an immediate child -- for each $p \in h$, a plan in the hypothesis, and for each $o$, open frontier item in $p$, if $o\in \Sigma$ it will try to directly replace $o$ with $\sigma_n$. This is performed by the function $\Call{CombineDirectly}{}$.
\item As a fragmented child -- for each $p \in h$, a plan in the hypothesis, and for each $f$, a fragment for $\sigma_n$, it will try to fuse $p$ with $f$. This is performed by the function $\Call{CombineAsChild}{}$.
\item As a sibling -- for each $t \in h$, a plan in the hypothesis, and for each $f$, a fragment for $\sigma_n$, it will try to find a rule $\alpha_0 \rightarrow \alpha_1, \ldots, \alpha_k : \phi$ in which both the root of $f$ and the root of $p$ are in $\alpha_1, \ldots, \alpha_k$. It will then create a higher level in the plan such that the new root will be $\alpha_0$, all the leaves but two will be open frontiers and the other two will be replaced with $p$ and $f$. This is performed by the function $\Call{CombineAsSibling}{}$. The fragments must obey the ordering constraints in $\phi$. The timestamp of the new fused fragment is the smallest timestamp of the original fragments.
\item Independently -- $h' = h \cup \{f | f$ is a fragment for $\sigma_n\}$. We must always consider the case that a new fragment is part of a new plan, or that it will be connected to another fragment in a later stage.  This is performed by the function $\Call{CombineDirectly}{}$.
\end{enumerate}

\begin{figure}[t]
 \centering
	\includegraphics[width=7cm]{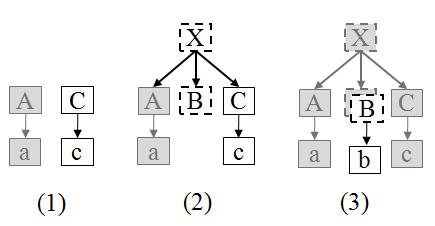}
 \caption{Example Hypotheses and Fusion Process}
        \label{fig:fusion}
 \end{figure}

Figure~\ref{fig:fusion} shows the process of constructing some of the possible hypotheses, using the previously defined plan library and the observation sequence $a,c,b$. For the first observation, $a$, SLIM will output a single hypothesis with the fragment for $A \rightarrow a$. Upon this hypothesis and the new observation $c$, two new hypotheses can be constructed: hypothesis (1) will be built using the $\Call{CombineIndependently}{}$ method and hypothesis (2) will be built using the $\Call{CombineAsSibling}{}$ method. Next, when trying to combine the third observation, $b$, it can be fused to hypotheses (2) with the $\Call{CombineAsChild}{}$ method, resulting with hypothesis (3). Hypothesis (3) shows the shadowed hypothesis after $a, c$ and the fusion with the new fragment for $b$. This is not the only valid hypothesis for these observations , both hypotheses (1) and (2) can be extended using the $\Call{CombineIndependently}{}$ method and the fragment for $B \rightarrow b$, thus adding two more possible hypotheses.

\textbf{Complexity Analysis} The bottleneck of both PHATT and SLIM is constructed from checking all combinations to add the new observation to the hypothesis set, which depends on the hypothesis set size from previous steps. 

Intuitively, although SLIM will create more hypotheses than PHATT (due to inability to enforce ordering constraints between fragments), the size of each hypothesis is smaller and the number of possible combinations will be smaller. To analyze the complexity, we define $C_i$ to be the size of options to combine the i-th observation. For PHATT, $C_i$ was analyzed to be $O((w_i\times b)^{h_i})$ where $h_i$ is the maximum depth of a plan (for recursive grammars, $h_i$ is bounded artificially), $w_i$ is number of open frontiers in all hypotheses and $b$ is the maximal OR branching factor of the plan library. 
Unlike PHATT, SLIM works bottom-up and develops a fragment instead of a complete path to the root and each fragment can be joined in a way that will increase either $w_i$ (by joining as a sibling or independently) or $h_i$ (by joining as a child) but not both. This means that for SLIM, the size of the combinations at each step will be $C_i=O((w_i\times b)^{log(h_i+w_i)})$.

%

\subsection {Top-Down Inference}
The generated hypotheses from SLIM's bottom-up parsing are local hypotheses that do not commit to the agent's underlying goal. 
However, we might wish to know the hierarchy of the outputted plans up until the goal. Usually, we don't require the complete set of hypotheses, just the k-most probable ones. In the tutoring scenario, for example, the complete plans might be sent to a teacher for review, or used to aggregate the paths used by different students. 

A similar difference between the goal-rooted and local hypotheses was defined in the SBR's paper as querying about current state vs. querying about history of states, where the ``current state'' query is similar to SLIM's bottom-up output and the ``history of states'' query is similar to PHATT's output.
We wish to be able to perform a ``history of states'' query in SLIM, meaning to query about plans that cover complete paths from the observed actions to goals. This is done by calling the PHATT algorithm with minor modifications. 

As discussed in previous sections, PHATT takes an observation and generates a complete path to a goal root. Here, we feed the PHATT algorithm with the fused fragments of a hypothesis instead of with observations. The rest of the process remains the same -- PHATT tries to combine the local plans to goal-rooted plans. We perform this for the k-most probable hypotheses.
In order to keep consistent with the ordering constraints, the fragments must be given to the modified PHATT by the order they were constructed, which is the timestamp that each fragment receives upon creation.
This process can be exemplified using hypothesis (a) from Figure~\ref{fig:fusion}: when requesting for goal-rooted plans, the modified PHATT will first receive the fragment for $A$, as it was created earlier, and will produce a plan like plan (1) in Figure~\ref{fig:example}. Next, it will receive the fragment for $C$, which it will be able to combine with plan (1).
Like in the example described above, there might be an overlap between the hypotheses constructed in the bottom-up step and the hypotheses constructed in the top-down step. In order to avoid such duplication, we must force the top-down inference to discard the leftmost trees which we already constructed. This is done by a straightforward bookkeeping, where we check all leftmost trees which were already produced by the bottom-up inference.
The complexity of this step is same as the complexity of PHATT.

\textbf{Completeness} 
The local hypotheses can be ranked by their probabilities, which is calculated as the multiplication of the constituting fragments' probabilities. Since the underlying probability model is the same, performing top-down inference for the k-most probable local hypotheses will result with the k-most probable hypotheses in PHATT, iff the prior probability of the goals distribute uniformly.

\vspace{-0.3cm}
\section{Empirical Evaluation}
We evaluated the SLIM algorithm on a simulated domain, based on AND/OR trees used by Kabanza et al.~\shortcite{kabanza2013controlling}. We used their same configuration which includes 100 instances with a fixed number (100) of
basic actions, five identified goals, a branching factor of 3 for AND rules
in the grammar and a branching factor of 2 for OR rules. Using these settings, each instance contains a sequence of 9 basic actions.

After compiling the rules to match the grammar representation in this paper, we got a plan library with 140 complex level actions and 244 rules.
We ran both PHATT and SLIM's bottom-up inference on the 100 instances and compared runtimes, number of created hypotheses and $C_i$ for each algorithm. 
We did not run SBR on this domain due to complexity issues -- SBR requires the rules to be modeled using only OR graphs. The transfer to this formalization requires $2^{24}$ different rules, which makes it unfeasible for this domain.

\begin{figure}[t]
 \centering
	\includegraphics[width=7cm]{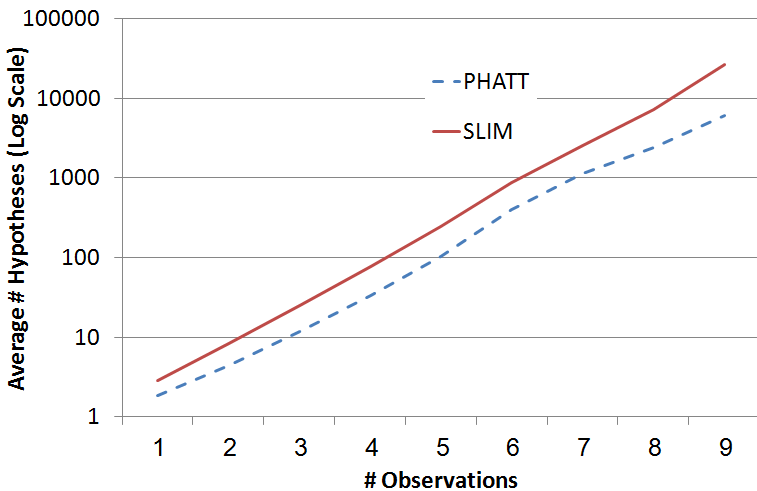}
 \caption{Average Number of Hypotheses per Length of Observations Sequence}
        \label{fig:emp1}
 \end{figure}

\begin{figure}[t]
 \centering
	\includegraphics[width=7cm]{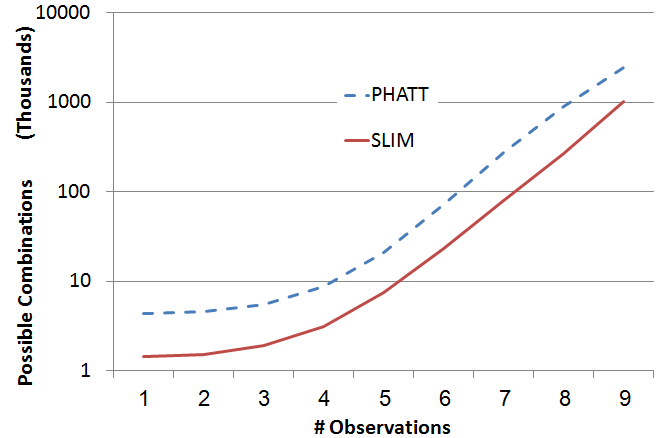}
 \caption{Average Value of  $C_i$ (Possible Combinations) per Length of Observations Sequence}
        \label{fig:emp2}
 \end{figure}


\begin{table}[t]
\centering
\begin{tabular}{|l|c|c|c|c|c|c|c|c|c|}
\hline
Obs.            &1	&2	& 3      & 4      & 5       & 6    & 7  	&8	&9  \\
\hline
PHATT           & 0.24 & 0.34 & 0.55 & 0.84 & 1.18 & 1.66 & 2.42 & 4.46 & 9.48 \\
SLIM  		& 0.16 & 0.16 & 0.18 & 0.25 & 0.43 & 0.74 & 1.21 & 1.92 & 4.45 \\
\hline
\end{tabular}
\caption{Average Runtime (in seconds) of SLIM's bottom-up inference and PHATT}
\label{tab:runtime}
\end{table}

As seen in Figure~\ref{fig:emp1}, the number of local hypotheses that the algorithm requires to keep is still exponential as in PHATT. SLIM even contains a larger number of hypotheses than PHATT, though not statistically significant.  However, since in PHATT a hypothesis contains goal-rooted plans right from the first observation, the depth of the plans is larger and there are more open frontiers to consider.  SLIM, on the other hand, contains plans as shallow as possible. 
The cost of PHATT's long hypotheses can be seen in Figure~\ref{fig:emp2}, where it is shown that the number of combinations SLIM requires to consider is an order of magnitude less than PHATT ($p \leq 0.05$). 

Table~\ref{tab:runtime} presents the runtime of the algorithms. These tests were conducted on the same commodity computer and were based on the same code, differing only in the step function of the different algorithms. Using these settings, SLIM was significantly faster the PHATT at every step ($p \leq 0.05$).

\begin{figure}[t]
 \centering
	\includegraphics[width=7cm]{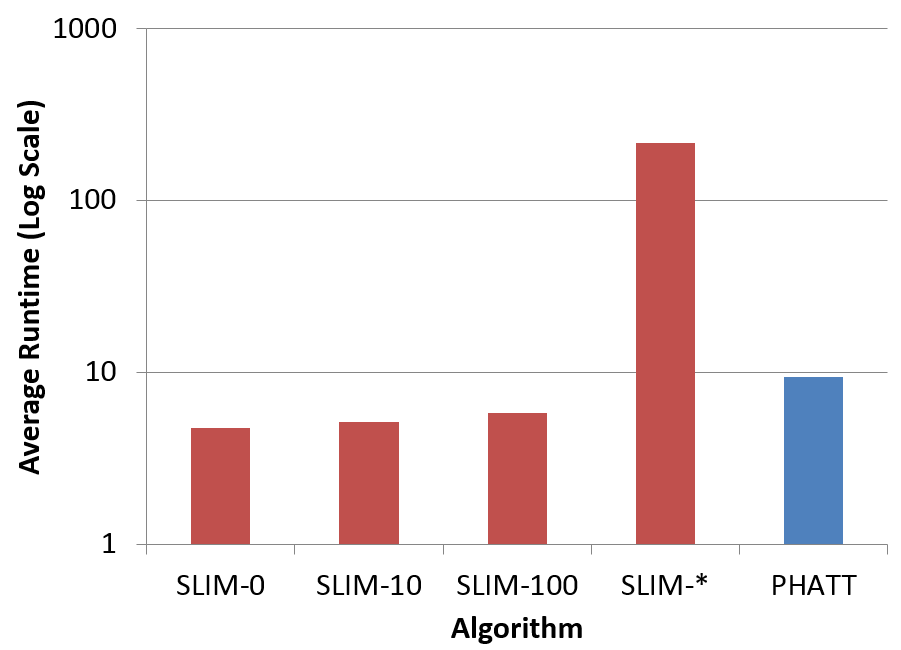}
 \caption{Average Runtime (in Seconds) per Algorithm Variant}
        \label{fig:emp3}
 \end{figure}

Next, we ran several variations of SLIM, notated using SLIM-i where i is the differing number of hypotheses calculated by the top-down inference (for example, calculating 0-most probable hypotheses, meaning only the bottom-up inference is performed, is notated SLIM-0).
We used these variations to evaluate the difference in runtime for the top-down step. The results are shown in Figure~\ref{fig:emp3}. SLIM-* is the complete compilation of all the hypotheses (which can be more than 10,000, as seen in Figure~\ref{fig:emp1}) to goal-rooted ones. As seen in this table, calculating the 100-th most probable hypotheses to the root was not significantly slower than SLIM-0 and still faster than PHATT ($p \leq 0.05$). Compiling the complete set of hypotheses will induce a significant increase in the runtime. However, 
for the challenge of creating a robust online plan recognition algorithm that can output description of the agent's actions in real-time, the results show significant improvement for SLIM in comparison to PHATT.

\vspace{-0.3cm}
\section{Discussion and Conclusion}
In this paper we presented a new plan recognition algorithm which uses the benefits of two well-known algorithms.
We showed that compared to state of the art algorithms, both in terms of expressiveness and capabilities, SLIM is superior.

The price SLIM's bottom-up inference pays to gain faster performance, is that the ultimate goals of the agents are no longer available. However, at any given point in time 
 the top-down inference can combine the set of fragments into goal-rooted hypotheses by looking at each fragment's root as an observation. 
 Moreover, in cases where the bottom-up inference manages to reach the goal, the top-down process may no longer be necessary. To conclude, for the purpose of reasoning about the agent's actions in real-time, it is sometimes sufficient to know only the complex actions that are already being executed, as SLIM's bottom-up inference outputs. If at any point the inference of the goals becomes important, the top-down inference complements the process, either by providing small set of hypotheses in real-time, or by inferring the complete set offline.

The most immediate investigation we wish to pursue is 
making SLIM an anytime algorithm that provides an increasing number of goal-rooted hypotheses. 
We also intend to improve the top-down step of the algorithm. 
by to using an existing goal recognition algorithm, which is significantly faster, and check consistency of the proposed goals with the outputted plans from SLIM's bottom-up mechanism.



\section*{Acknowledgments}
This work was supported in part by EU FP7 FET project no. 600854, and the Israeli Science Foundation Research Grant no. 1276/12. R.M. is a recipient of the Pratt fellowship at the Ben-Gurion University of the Negev.
\bibliography{libb}
\bibliographystyle{named}

\end{document}